\documentclass[10pt,twocolumn,letterpaper]{article}

\usepackage{cvpr}              % To produce the CAMERA-READY version
\usepackage{graphicx}
\usepackage{amsmath}
\usepackage{amssymb}
\usepackage{booktabs}

\usepackage[pagebackref,breaklinks,colorlinks]{hyperref}

\usepackage[capitalize]{cleveref}
\crefname{section}{Sec.}{Secs.}
\Crefname{section}{Section}{Sections}
\Crefname{table}{Table}{Tables}
\crefname{table}{Tab.}{Tabs.}

\def\cvprPaperID{*****} % *** Enter the CVPR Paper ID here
\def\confName{CVPR}
\def\confYear{2023}

\begin{document}

%%%%%%%%% TITLE - PLEASE UPDATE
\title{Human Action Recognition in Egocentric Perspective Using 2D Object and Hands Pose }

\author{Wiktor Mucha\\
Computer Vision Lab, TU Wien\\
Vienna, Austria\\
{\tt\small wiktor.mucha@tuwien.ac.at}
% For a paper whose authors are all at the same institution,
% omit the following lines up until the closing ``}''.
% Additional authors and addresses can be added with ``\and'',
% just like the second author.
% To save space, use either the email address or home page, not both
\and
Martin Kampel\\
Computer Vision Lab, TU Wien\\
Vienna, Austria\\
{\tt\small martin.kampel@tuwien.ac.at}
}
\maketitle

%%%%%%%%% ABSTRACT
% \begin{abstract}
%    The ABSTRACT is to be in fully justified italicized text, at the top of the left-hand column, below the author and affiliation information.
%    Use the word ``Abstract'' as the title, in 12-point Times, boldface type, centered relative to the column, initially capitalized.
%    The abstract is to be in 10-point, single-spaced type.
%    Leave two blank lines after the Abstract, then begin the main text.
%    Look at previous CVPR abstracts to get a feel for style and length.
% \end{abstract}

%%%%%%%%% BODY TEXT

\section{Introduction}

% 1.Subject or Problem addressed in the paper

Egocentric action recognition popularity has grown significantly in recent years as the quality of wearable cameras improves and this technology promises numerous potential applications such as healthcare and Active Assisted Living (AAL). The combination of the low costs of the technology and the ease of use results in the rising number of available datasets for the research community. This study places the focus on egocentric action recognition employing hand and objects 2D pose for sequence classification using a state-of-the-art Transformer based method. Most of the literature focuses on exploring 3D hand pose information for action recognition \cite{tekin2019h+, das2021symmetric, Kwon_2021_ICCV}. However, these studies do not employ sensor-acquired depth information but generate its estimation from RGB frames, which results in error in pose prediction oscillating around mean End-Point Error (EPE) equal to 37 mm \cite{Kwon_2021_ICCV}. Exploiting 2D skeleton data is a promising approach for hand-based action classification. It demands less computational power and offers privacy enhancement over processing complete RGB frames, which might contain sensitive data raising concerns. The experiment described in this study aims to answer how feasible 2D hand and object pose information is for action recognition tasks compared to current state-of-the-art methods, including 3D-based methods. In the first stage of our study, we utilise hand and object poses provided by the \textit{H2O Dataset} \cite{Kwon_2021_ICCV} and transform them into the 2D space. Our method achieves validation results of 94\% outperforming other existing solutions. The accuracy of the test subset drops to 76\%, highlighting the essential generalisation improvement. The second stage is planned to employ its own methods for pose estimation and shows promising results.

\section{Related Work}

\begin{figure*}[th]
\begin{center}
\includegraphics[width=0.72\textwidth]{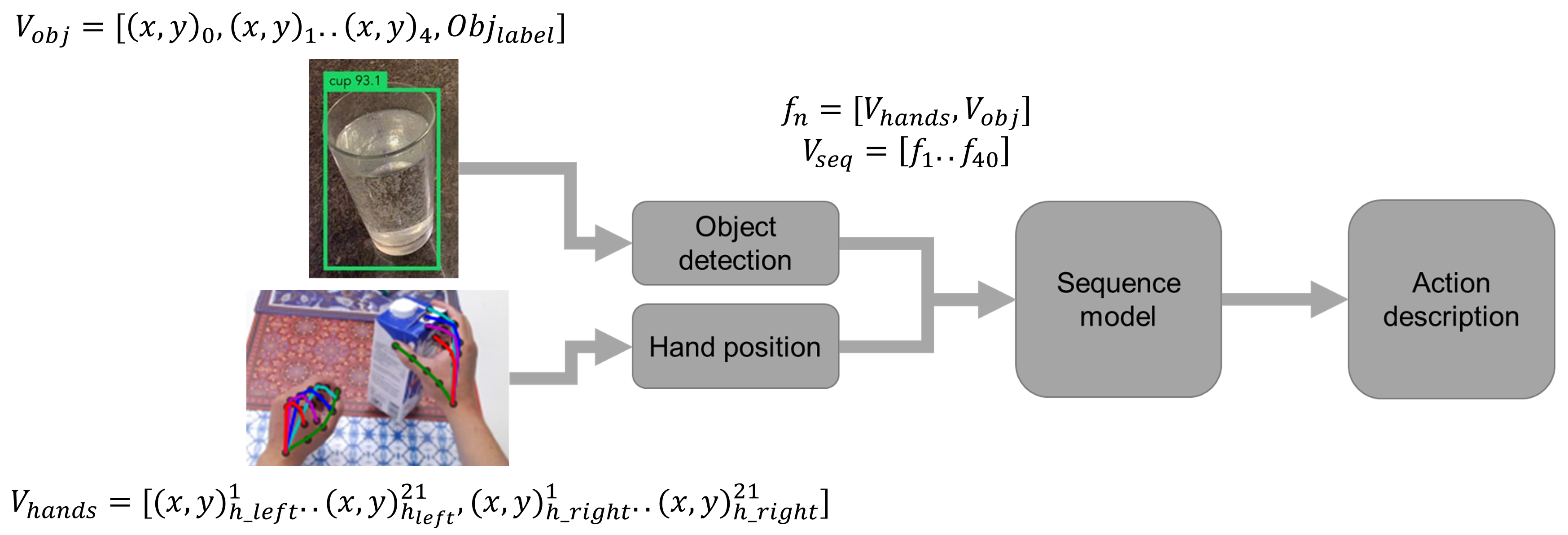}
\vspace{-0.6cm}
\end{center}
 \caption{ \textbf{Visualisation} of the entire procedure for action recognition.}
\label{ego_system}
\end{figure*}

Egocentric action recognition is tackled through various approaches. One recurring strategy involves employing hand and object information for action description. Across scientific literature, we observe instances of utilising 2D data as input in the recognition process, like the work of Cartas et al. \cite{cartas2017contextually} who propose the employment of object detector based on a Convolutional Neural Network (CNN) to assess primary region (hands) and secondary region (objects) positions. Sequences of these positions are further processed using a Long Short-Term Memory (LSTM) network. Nguyen et al. \cite{nguyen2019neural} go beyond bounding box information and exploit 2D hand skeleton information obtained with a CNN. Estimated joints are aggregated using spatial and temporal Gaussian aggregation. In order to perform action classification, the network learns the SPD matrix.

% 3D
Contrary to studies investigating 2D data, other authors researching hand role in action recognition focus on utilising 3D information\cite{tekin2019h+, das2021symmetric, Kwon_2021_ICCV}. Tekin et al. \cite{tekin2019h+} uses a single RGB frame to estimate 3D hand pose and object pose. For this purpose, they use a CNN architecture and LSTM network to classify actions. Das et al. \cite{das2021symmetric} use an architecture based on the spatiotemporal graph CNN to recognise the action. In their method, they create subgraphs for a movement of the description of each finger. Kwon et al. \cite{Kwon_2021_ICCV} create subgraphs for each hand and object that are later merged into multigraph representation. This approach allows to learn about interactions between each of these three.

In the mentioned studies, the 3D hand pose is regressed from 2D to 3D using neural networks and intrinsic camera parameters \cite{tekin2019h+, Kwon_2021_ICCV}. These methods provide promising 3D results but do not employ sensor-acquired depth maps like a stereo camera or a depth sensor. The mean EPE acquired on \textit{H2O Dataset} \cite{Kwon_2021_ICCV} oscillates around 37mm, which is far from accurate considering human hand size. A potential solution is depth sensor utilisation which except for accurate 3D information enhances privacy \cite{mucha2022addressing}, but the market lacks wearable depth sensors.
In contrast, the proposed method based on 2D data permits generalisation to any dataset as it does not require intrinsic camera parameters. Such a network can also have fewer parameters, causing computation to be less demanding than architectures generating 3D data. Finally, results on popular egocentric benchmarks like \textit{H2O Dataset} or \textit{Epic-Kitchen} \cite{Damen2018EPICKITCHENS} are missing.

\section{Methodology}

% \begin{figure}[t]
%   \centering
%   \includegraphics[width=\linewidth]{cvpr2023-author_kit-v1_1-1/latex/method.png}
%   \caption{
%   Visualisation of the entire procedure for action recognition.
%   % Visualisation of our entire procedure for hand pose estimation in the egocentric domain.
%   }
%   \label{ego_system}
% \end{figure}

The proposed action recognition model is constructed from three separate modules. First, the hand pose network estimates the position of hands in 2D space using 21 key points for each hand. A separate block is responsible for locating the object's position. 
Each hand pose is represented by $\mathit{Ph}_{t}^{i}(x,y)$ where 
$\mathit{t} \in \{l,r\}$ describes the left or right hand and $\mathit i \in [1..21]$ stands for the hand keypoint number. Respectively, the object is represented as $\mathit{Po}_{bb}^{i}(x,y)$ where $\mathit i\in[1..4]$ corresponds to the bounding box corners and $\mathit{Po}_{l}$ is the object label. Frames $\mathit f_n$ are a concatenation of these points creating the input vector $\mathit V_{seq}$ described as:
\begin{equation}
f_n = Ph_{l}^{i}(x,y) \oplus Ph_{r}^{i}(x,y) \oplus Po_{bb}^{i}(x,y) \oplus Po_{l}
\end{equation}
\begin{equation}
V_{seq} = [f_0..f_n], n \in [1..40]
\end{equation}
This information vector $\mathit V_{seq}$ is passed to the last stage, which is the sequence model. In the experiment, we employed a model inspired by Visual Transformer by Dosovitskiy et al. \cite{dosovitskiy2020image}. The method is depicted in Figure \ref{ego_system}.

To embed each frame into $\mathit f_n$ representation, the primary step is object detection with YOLOv7 \cite{wang2022yolov7} fine-tuned to detect objects, left and right hand locations in the image. Further, a single-hand pose model estimates hand skeletons for the provided regions from the previous phase. This architecture is trained on \textit{FreiHAND} dataset \cite{zimmermann2019freihand} and it is built from the \textit{EfficintNetV2} \cite{tan2021efficientnetv2} feature extractor.
% and prediction head inspired by the work of Xiao et al. \cite{xiao2018simple}.

\section{Evaluation}

Our experiment employs \textit{H2O Dataset} for egocentric action recognition, which provides Ground Truth (GT) action labels for 36 actions, hand pose information and object positions. The study is planned in two phases. The first stage aims to determine the robustness of the proposed method based only on 2D hand pose and object pose only. The second stage is currently in progress, and it is planned to include a complete model estimating its own 2D hand and object poses. We find that our method works best when utilising 40 frames for a sequence randomly sampled during training and equally sampled for the testing phase. Data augmentation is performed by horizontal flipping and random cropping of the sequences.

\begin{table}[t]
\caption{\textbf{Results of various action recognition methods} performing on \textit{H2O dataset} expressed in accuracy.}
\label{tab:results}
\begin{tabular}{@{}lcc@{}}
\toprule
Method:       & Validation {[}\%{]} & Test {[}\%{]} \\ \midrule
I3D \cite{carreira2017quo}          & 85.15                        & 75.21                  \\
SlowFast \cite{feichtenhofer2019slowfast}     & 86.00                        & 77.69                  \\
H+O \cite{tekin2019h+}          & 80.49                        & 68.88                  \\
ST-GCN \cite{yan2018spatial}    & 83.47                        & 73.86                  \\
TA-GCN \cite{Kwon_2021_ICCV}       & 86.78                        & \textbf{79.25}         \\ \midrule
Ours-HandP+ObjL & 73.77               & 56.19                  \\
Ours-HandPL+ObjP+ObjL & 80.33              &   62.80                \\
Ours-HandPR+ObjP+ObjL & 81.97              & 67.35                  \\
\textbf{Ours-HandP+ObjP+ObjL} & \textbf{94.26}               & 76.03                  \\ \bottomrule
\end{tabular}
\end{table}

The first stage experiment is based on the given GT. The 3D points describing each hand pose and 6D object poses are transformed into a 2D plane using the given camera intrinsic parameters.
The results of our action recognition method on the validation subset are optimistic, as they outperform other methods obtaining 94.26\% and are presented in Table \ref{tab:results} as \textit{Ours-HandP+ObjP+ObjL}. 
The accuracy on the test subset drops to 76.03\%, placing it among other methods. These preliminary results highlight the difference between validation and test datasets which should be addressed in the future by improving data augmentation but show promising accuracy of the 2D-based method.

\begin{figure}[t]
  \centering
  \includegraphics[width=\linewidth]{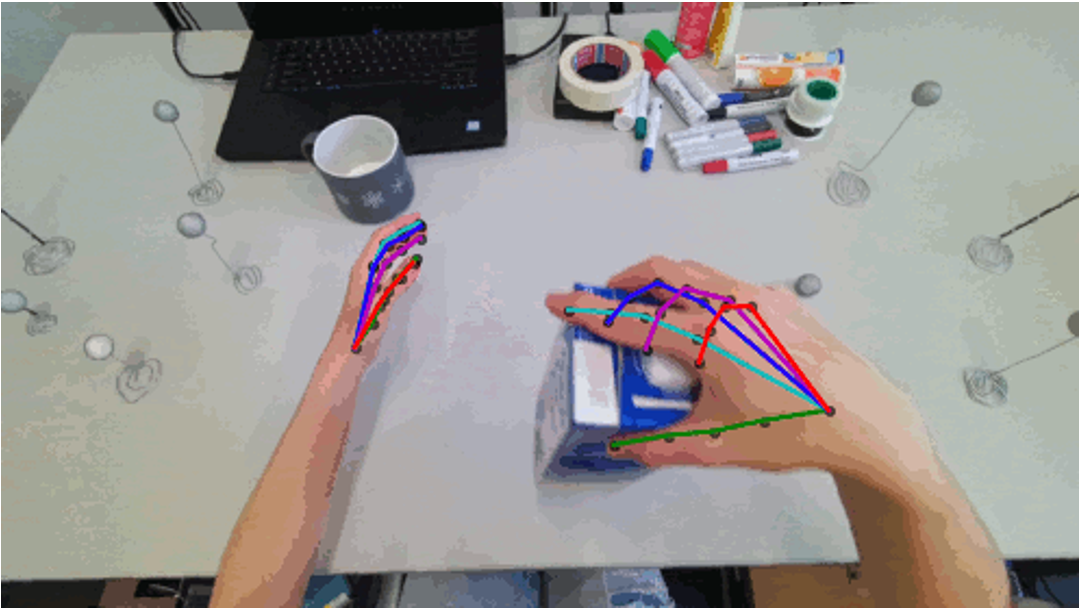}
  \caption{
  \textbf{Example of action picking up a milk box} from \textit{H2O Dataset} using our method for 2D hand pose description.
  % Visualisation of our entire procedure for hand pose estimation in the egocentric domain.
  }
  \label{action_example}
\end{figure}

Evaluation of the second stage is in progress. Fine-tuning of the YOLOv7 model results in 0.995 mAP@0.5, causing the object information to be accurate for further inference. The hand pose estimation part is more challenging due to occlusions caused by objects or self-occlusion. Our method for hand pose estimation misses only 14 hand detections in 55747 training frames, and results in EPE equal to 24.46 pixels in a 1280x720 input image. The example frame from \textit{H2O Dataset} with pose estimating using our method is presented in Figure \ref{action_example}.

\subsection{Ablation study}
To determine the importance of each module for the final accuracy of action recognition an ablations study is performed. Scenario using only the object labels and hand poses, skipping bounding box information results in the drop of accuracy to 73.77\% validation and 56.19\% test. Results are presented in Table \ref{tab:results} as \textit{Ours-HandP+ObjL}. Further ablations focused on the role of each hand in the scene. For this purpose, the vector of one of the hands was exchanged for zero values. \textit{Ours-HandPL+ObjB+ObjL} and \textit{Ours-HandPR+ObjB+ObjL} represent, respectively, the usage of only the left and only right hand for action recognition. These results show the importance of including information about the manipulated object and the presence of at least one hand manipulating it. 

\section{Conclusion}

The study implemented an egocentric action recognition method based on hand and object 2D pose information. This position information is used to describe each frame of a sequence and is embedded for sequence classification using a transformer-based network. At this stage, the preliminary study conducted with given positions yielded promising results, outperforming other methods' accuracy on validation data and performing closely on test data.

%%%%%%%%% REFERENCES
{\small
\bibliographystyle{ieee_fullname}
\bibliography{camera_ready}
}

\end{document}